# Diversity-aware social robots meet people: beyond context-aware embodied AI

*Carmine Recchiuto, Antonio Sgorbissa*

**Introduction**

Mayra is a 34-year-old woman from Sri Lanka who arrived in Genoa in 2020, just before the COVID-19 pandemic. Mayra spends her days taking care of her three children and doing housework. Due to the lockdown measures, she had few opportunities to develop relationships with Italian people, so her Italian has remained very basic. The situation did not improve until December 2021 because finding a job was challenging due to the remaining COVID restriction. In January 2022, her husband bought a small robot, and Mayra called it "Dhvija." Mayra adores it; she says the robot knows about Sri Lanka and is very respectful. It helps her learn Italian and things about Italy. Yesterday, Dhvija suggested that Mayra check out a local organization where women meet and share their cooking experiences. Today, Mayra taught her new friend Alessandra how to prepare dhal curry. "I need to ask Dhvija where to buy ingredients for pesto," she thinks, smiling.

Michele is a 72-year-old man who lives in Padua. He uses a wheelchair because of an accident that severely affected the mobility of his legs and left arm. He says he does not feel "confined to the wheelchair" because "the wheelchair is not a prison" but helps him move about and be independent. His daughter is worried because he is getting older and lives alone. So she rented a robot called Doc to help him with easy tasks and keep an eye on him while at home. Michele is amazed at how smart the robot is: Doc knows that his left arm is paralyzed, so it always approaches him on his right side when it hands him an object. Also, it knows how to move around the house, leaving space for the wheelchair. Michele is satisfied with this new tool. "But how can it be," he asks himself, "that this appliance knows me so well?".

Andrea is a 34-year-old non-binary person who lives in Bergamo. Andrea often goes shopping at the local supermarket, which recently bought a social robot to welcome customers named Wise. Andrea is surprised at how respectful Wise is when talking to people. Wise asked Andrea how to address them, and unlike some people, the robot did not seem confused. Also, the robot uses gender-inclusive pronouns and adjectives, which is very complex in Italian. "Why are humans not as wise as Wise"? Andrea wonders, laughing.

The scenarios above introduce a key concept: how challenges related to health and social care may be addressed through socially assistive robots which not only support independent, active and healthy lifestyles and provide personalized advice and guidance but are also able to value diversity, promote inclusion and equal opportunities to enable a successful and rewarding experience.

**State-of-the-art**

In recent years, starting from research in Transcultural Nursing [1] and culturally competent Health Care [2, 3], the EU-Japan project CARESSES [4, 5] addressed for the first time the problem of developing culturally competent Socially Assistive Robots for elderly care. The concept of "culture" is complex [6, 7, 1, 8, 9, 10], and there is no consensus among researchers on how to define it [11, 12]. Nevertheless, a simple yet effective definition holds that culture is a shared representation of the world of a group of people: by "culturally competent," we mean a robot that can adapt its perceptions, plans, actions, and interaction style depending on the worldview of the person it is interacting with, including their beliefs, values, ideas, language, norms and visibly expressed forms such as customs, art, music, clothing, food, etc. [13, 14, 15, 16].

We think it is now crucial expanding this concept: robots should be more than culture-aware or culturally competent: they should be able to take diversity into account in all possible dimensions. According to a broadly accepted definition, "diversity is about what makes each of us unique and includes our backgrounds, personality, life experiences and beliefs, all of the things that make us who we are. (…) Diversity is also about recognising, respecting and valuing differences based on ethnicity, gender, age, race, religion, disability and sexual orientation. (…) Inclusion occurs when people feel, and are valued and respected regardless of their personal characteristic or circumstance (…) Equal opportunity means that every person can participate freely and equally in areas of public life (…) without disadvantage or less favourable treatment due to their unique attributes." [17]

Diversity awareness will introduce a crucial innovation in social robotics. It will produce robots that can re-configure their behavior to recognize and value the uniqueness of the person they interact with to promote respect for diversity, inclusion, and equal opportunities, not only for health and social assistance [18, 19] but also cooperation in the workplace [20], education [21], welcoming visitors [22].

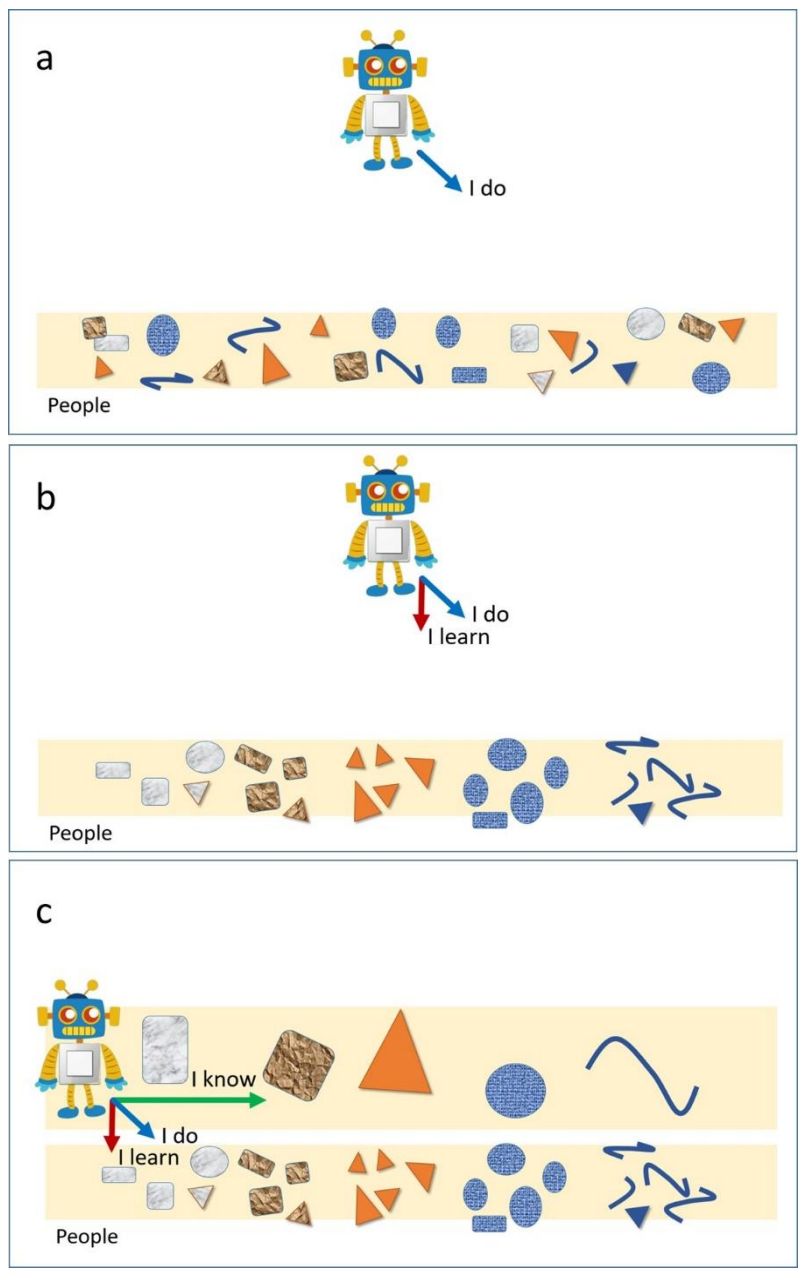

*Figure 1 a) a robot that needs to interact with diverse target groups and people and do things in a way that maximizes the chance of success in a statistical sense; b) the typical approach that tries to learn the characteristics of people during the interaction to improve performance; c) the diversity-aware approach, that uses a prior that uses a priori knowledge about the shared needs and preferences of different target groups as a starting point to learn faster and better.*

From a practical perspective, the problem of "customizing a robot" to match a person's profile has been addressed in AI and Robotics [23]. Ideally, roboticists plan to address this problem through Machine Learning techniques to the end of acquiring new knowledge about the person and its environment by periodically collecting and processing a vast amount of data, possibly through continual lifelong learning [24]. However, even if ML has produced incredible results in many different domains, it is hard to imagine a system capable of learning "everything" about the person only through data collecting and processing. Indeed, using ML tools requires us to accurately choose the most appropriate models depending on what we want the robot to learn: no "general purpose" learning solution exists as like as no "strong AI" [25] exists, i.e., a hypothetical machine with the ability to apply intelligence to any problem, rather than just one specific problem.

Then, as long as we have a priori knowledge about the preference and needs of diverse groups and people, why not use this knowledge? Figure 1 summarizes this concept: on the top, a robot that needs to interact with diverse target groups and people and do things in a way that maximizes the chance of success in a statistical sense; in the middle, the typical approach that tries to learn the characteristics of people during the interaction; on the bottom, the diversity-aware approach, that uses a priori knowledge about the shared needs and preferences of different target groups as a starting point to learn faster and better the individual characteristic of everybody – that, of course, are different within each group.

The general problem of making intelligent systems aware of differences has been addressed before. For instance, the negative impact of biases has recently come under the spotlight due to so-called "racist" algorithms [26, 27, 28], which programmers unconsciously design or train using their own (presumed neutral but actually biased) models of the world, thus producing outputs that are prone to errors and may be unfair [29, 30, 31]. The recent field of FAIR ML addresses these problems and shares some similarities with diversity-awareness: indeed, they both aim to give people equal opportunities by preventing sensitive attributes (e.g., ethnicity, sex, religion, age, pregnancy, familial, or disability status) to be used for discrimination (e.g., concerning healthcare, employment, education, housing, credit). However, FAIRness emphasizes ML for perception and decision making: diversity-aware robots, instead, shall be capable of real-time HRI by considering diversity as a key feature in all aspects, including sensor data interpretation, knowledge representation, action planning and execution, verbal and non-verbal communication.

**A framework for diversity-awareness.**

Since diversity-awareness is closely related to culture-awareness, as ethnicity and culture are two dimensions of diversity that inclusive robots shall consider, we think the problem of diversity awareness can start from previous attempts to design culture-aware robots.

No earlier approach (except CARESSES [15]) aimed to define a general conceptual framework representing cultural knowledge for intelligent systems. In his seminal work and later studies [7, 32], Hofstede introduced the Dimensions of Culture[1], a quantitative representation of values and practices found in human cultures to measure the similarities and differences across cultures. AI and Robotics researchers often use Hofstede's model as a reference because of its computational nature. Still, this model is insufficient to produce culturally competent behavior: different cultures imply conceptualizations of the world primarily symbolic. Therefore, they cannot be summarized by five (or more [32]) numerical indices. For example, what is a wedding in different cultures? A wedding ceremony? A wedding party? When hearing these words, which images appear in the mind of an Italian, Greek, Indian, or Swedish person? How do these worldviews differ from each other?

Along this line, in CARESSES [15], we proposed a formal, machine-processable model of cultural knowledge that takes inspiration from the "ontological turn" [10] that has produced an intense debate in anthropology in the last decade [12]. Please notice that, in summarizing, this debate turns around the question of whether cultures should be considered as different representations of the "real world" (i.e., worldviews) or rather multiple worlds. The debate is ideally re-composed with "computer Ontologies" since they are undoubtedly a 'representation,' either you consider them as "worlds" or "worldviews," expressed through a formal (machine-processable) language. Specifically, the model we proposed is based on two core components:

---

[1] Power distance, Individualism vs. collectivism, Uncertainty avoidance, Masculinity vs. femininity, Long-term orientation vs. short-term orientation.

- **an Ontology** encoded in Description Logics and implemented in OWL 2 [33]), which stores a rich vocabulary (also referred to as Terminological Box to represent all the relevant concepts and relations for all the cultures considered (including beliefs, values, practices, customs, traditions, and systems' functionalities), as well as all the multiple worlds/worldviews corresponding to different cultures through proper instances and property assertions (also referred to as Assertional Box);
- **a probability distribution** over all cultures in the ontology described through a Bayesian Network [34] whose hierarchical structure mimics the ontology. The Bayesian Network relates assertions that hold for a cultural group with assertions that may hold or not hold for an individual self-identifying with that culture and is updated whenever new evidence is acquired through dialogue, observation, or other sources of information.

Each culture/world/worldview is modeled by instantiating all concepts and relations in the vocabulary and assigning them a probability. In our model, two cultures/worlds/worldviews do not differ in their concepts and relations. A concept that does not exist or is not relevant in a given culture is associated with a null (or very low) probability; emerging or disappearing concepts are modeled as permanent in the ontology but with a probability that dynamically changes with time [15].

When coming to diversity awareness along different dimensions such as gender, age, disability, and sexual orientation, the same model can be straightforwardly extended to all these new dimensions, Figure 2. Instead of having multiple world/worldviews corresponding to cultures, we can use the same computational model to represent multiple world/worldviews for all diversity dimensions we want to consider in our application domain, e.g., gender, age, cognitive capabilities, sexual orientation, etc. Moreover, as it happened for culture, this will be interpreted in a probabilistic sense to avoid stereotyping, making it possible for a person to have different needs and preferences than the target group they belong to.

When interacting with the person, the robot will only need to get the relevant knowledge that, in a probabilistic sense, is expected to be the most appropriate for the person, use it in the interaction, and revise knowledge as required using the feedback received during the interaction by updating probabilities as needed.

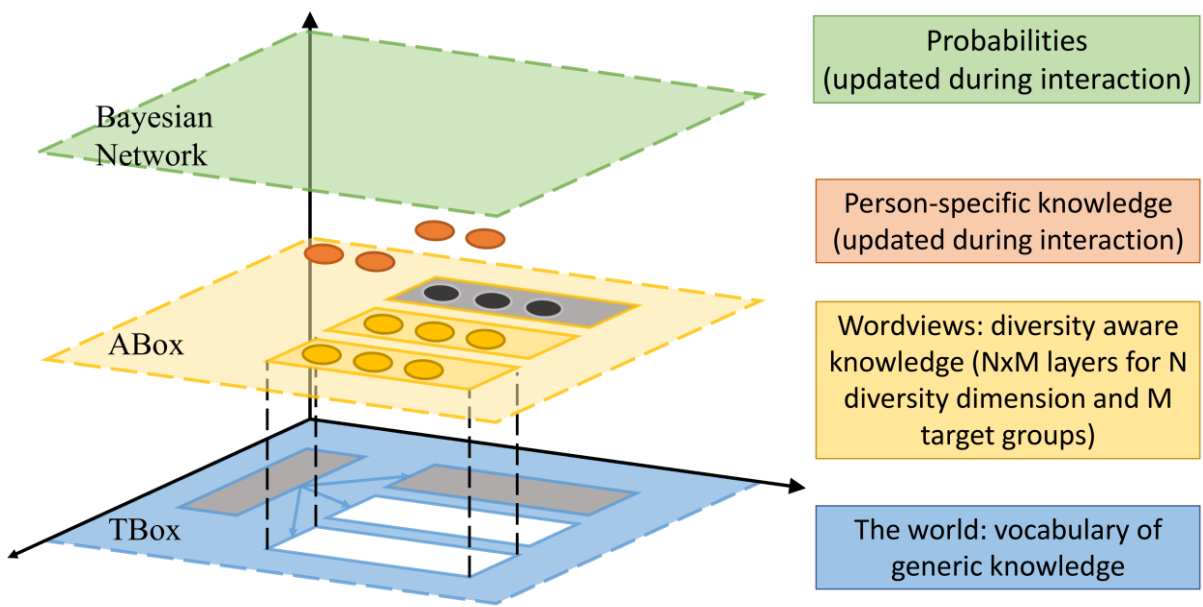

*Figure 2 Visual representation of a possible computational model for diversity-aware robots.*

**Conclusion**

Starting from our previous work with culturally-competent robots, we discussed the importance of developing diversity-aware robots: robots capable of valuing diversity, promoting inclusion, and equal opportunities to enable a successful and rewarding experience. Unfortunately, the problem has never been addressed in robotics. However, we claimed that developing a general framework for diversity-aware robots is an urgent need if we want robots to be accepted by people in society.

To this end, we discussed a possible technological solution to implement such a framework starting from popular tools in Artificial Intelligence, showing how the computational models that have already proved effective for culture-aware robotics can be easily extended to promote diversity awareness in intelligent systems.

**Acknowledgments**

The authors wish to thank Nicola Bellotto, Alberto Pretto, Monica Pivetti, Silvia Debattista, Linda Battistuzzi, Lucrezia Grassi for helping develop the concept of diversity-aware robotics.

Presented at the Roundtable "AI in holistic care and healing practices: the caring encounter beyond COVID-19", Anthropology, AI and the Future of Human Society, 6-10 June 2022, Royal Anthropological Institute

**Carmine Tommaso Recchiuto**, PhD., 38, is Assistant Professor at the University of Genova, where he teaches Experimental Robotics, ROS programming, and Computer Science. His research interests include Humanoid and Social Robotics (with a specific focus on knowledge representation and human-robot interaction), wearable sensors, and Aerial Robotics. He has been the Coordinator of software integration and Head of Software Development for WP2 in the CARESSES project, aimed at endowing social robots for older adults with cultural competence. He has also been the local Coordinator and Leader of WP2 for the BrainHuRo project, developing Brain-Computer Interfaces for humanoid robots' remote control. He is the author of more than 40 scientific papers published in International Journals and conference proceedings.

**Antonio Sgorbissa**, PhD, 52, is an Associate Professor at the Università degli Studi of Genova and a teacher in EMARO+ and JEMARO+, the European and Japanese Masters' in Advanced Robotics. He has 20 years of experience teaching ICT-related topics to students of non-technical universities in a simple, accessible, and appealing way. He has been coordinator of H2020 CARESSES (Culturally Aware Robots and Environmental Sensor Systems for Elderly Support) and Principal Investigator in National and EU projects, including DIONISO (a multidisciplinary effort focusing on ICT for intervention in earthquakes) and WearAmI (focusing on assistive robotics in smart environments). He is Associate Editor of the International Journal of Social Robotics, Journal of Advanced Robotic Systems, and Intelligent Service Robotics. He is a member of the board of directors and executive vice president of I-RIM, Italian Institute of Robotics and Intelligent Machines, Director of the Social Robotics Lab at University of Genova and RASES, the inter-university center on Robotics and Autonomous Systems in Emergency Scenarios. He is the author of more than 150 scientific articles indexed in international databases and has been awarded five patents as an inventor.